\documentclass[cameraready]{Interspeech}
\usepackage[flushleft]{threeparttable}
\usepackage{tablefootnote}
\usepackage{tipa}
\usepackage{algorithm}
\usepackage{algpseudocode}
\usepackage{amsmath}
\usepackage{placeins}
\usepackage{makecell}
\usepackage{amssymb}

\title{Balalaika: Data-Centric, Prosody-Aware Annotation Pipeline for Russian Speech}

\author[affiliation={1,2 }, orcid=0009-0001-8203-1059, equalcontribution, correspondingauthor]{Kirill}{Borodin}
\author[affiliation={1}, orcid=0009-0002-5559-470X, equalcontribution]{Nikita}{Vasiliev}
\author[affiliation={1, 2}, orcid=0009-0001-9935-7514]{Vasiliy}{Kudryavtsev}
\author[affiliation={1}, orcid=0009-0004-7716-8714]{Maxim}{Maslov}
\author[affiliation={1}, orcid=0000-0003-1739-9831]{Mikhail}{Gorodnichev}
\author[affiliation={1}, orcid=0000-0002-5802-5513]{Grach}{Mkrtchian}

\address{
    $^1$ lab260, Moscow Technical University of Communications and Informatics, Russia\\
    $^2$ BitmanagerAI, UAE
}

\email{kborodin.research@gmail.com}


\usepackage{comment}

\begin{document}

\maketitle

\begin{abstract}
    We introduce Balalaika, an open-source, data-centric pipeline for processing audio and producing prosody-aware annotations. It combines semantic VAD for context-preserving segmentation, multi-ASR ensembling with ROVER consensus decoding, while retaining word-level timestamps, followed by automatic quality and speaker-purity filtering. The text is further enriched with punctuation restoration, lexical stress and e/yo normalization, and IPA phonemes. Using Balalaika, we build a 5.1k-hour multi-source Russian corpus with rich annotations, and show consistent gains under equalized training budgets for both speech denoising and TTS; ablations confirm complementary benefits of stress and punctuation and improved synthesis with stricter MOS filtering.
    
    The dataset is publicly available at https://hf.co/collections/lab260/balalaika-dataset
    
    The full version paper is available at https://arxiv.org/abs/2507.13563
\end{abstract}

\section{Introduction}\label{sec:introduction}

The rapid growth of web audio~\cite{EdisonResearch2025} and voice interfaces~\cite{SNSInsider2025} makes scalable mining of speech content one of the core  challenges. As these technologies become integral to daily interaction~\cite{HOMBECK2024104069, voice_llm_interface}, they necessitate richly annotated, high-quality audio to ensure intelligibility and naturalness~\cite{GapBodhitaru2024, ATLTranslate2023}.

However, significant gaps hinder effective use of web audio for advanced applications: dataset creation still leans on manual curation~\cite{karpov21_interspeech, ardila2020commonvoicemassivelymultilingualspeech, zen19_interspeech} and audiobooks~\cite{celeste2019mailabs, ruls_slr96, sova2022dataset}, which is especially limiting for under-resourced languages with scarce high-quality data and tooling~\cite{karpov21_interspeech, Marren2012}. Russian presents unique complexities~\cite{rodionova2001word, 10.1162/tacl_a_00251} (vowel reduction~\cite{RUDN:TLSS34176}, palatalization~\cite{RUDN:TLSS34176}, and mobile stress~\cite{petrov-2025-ruaccent} affecting meaning and prosody) that prevailing methods often overlook. As a result, pipelines optimized for high-resource languages  tend to ignore phonetic and prosodic nuance, critical for morphologically rich languages~\cite{Saif2024, Marren2012}.

Gaps in scalable annotation and dataset frameworks critically undermine speech technologies, particularly in multilingual contexts. Simplistic annotation limits TTS quality~\cite{giraldo24_iberspeech, koizumi23_interspeech}, producing unnatural outputs due to poor handling of stress~\cite{petrov-2025-ruaccent} and phonetic nuances~\cite{RUDN:TLSS34176}. Furthermore, reliance on audiobooks introduces rigid, dictated patterns that lack spontaneous prosody~\cite{zhang23_ssw, Pethe2025}, ultimately degrading the performance and accuracy of generative speech tasks~\cite{peng2025surveyspeechlargelanguage}.

Beyond technical limitations, these gaps create barriers to multilingual  accessibility~\cite{zee-etal-2024-group}, excluding vast audio resources from  integration into web indexes~\cite{Saif2024}, which perpetuates biases toward high-resource languages~\cite{FENG2024101567}. They also contribute to stagnation in AI research, as models trained on inadequately annotated data underperform in speech processing tasks~\cite{lau2025dataqualityissuesmultilingual}, ultimately hindering equitable access to speech applications.

To address these challenges, we introduce Balalaika, a scalable framework for processing Russian audio, enabling high-quality datasets that capture  prosody and linguistic complexities for  generative models. This open-source methodology is novel in bridging gaps for Russian, as demonstrated by our creation of the dataset comprising over 5000 hours of high-quality speech. It supports audio mining through a scalable annotation pipeline, mitigating language biases in dataset creation.

To guide our investigation, we address the following research questions (RQs), which explore scalable annotation pipeline and its impact on generative speech models:

\textbf{RQ1 (Framework Comparison):} How does our annotation framework compare to existing methods in producing high-quality datasets for Russian speech?

\textbf{RQ2 (Denoising Performance):} How does training speech denoising models from scratch on our annotated data, under equal conditions, compare to training on other datasets?

\textbf{RQ3 (TTS performance):} How does training text-to-speech (TTS) models from scratch on our annotated data, under equal conditions, compare to training on other datasets in terms of naturalness and intelligibility?


To investigate these RQs, we built an open-source, modular pipeline that combines semantic VAD~\cite{smartturnv2} for segmentation, multi-ASR ROVER fusion~\cite{salute2024gigaam, vosk, ttech_tone_2025} with word-level timestamps, NISQA-based quality~\cite{Mittag_Naderi_Chehadi_Möller_2021} and pyannote speaker filtering~\cite{Plaquet23, Bredin23}, and prosody-aware text enrichment (punctuation~\cite{RUPunct}, lexical stress~\cite{petrov-2025-ruaccent}, and G2P phonemes); Section~\ref{sec:framework} details each stage.

Our experiments reveal that models trained on data obtained using our proposed Balalaika framework achieve higher scores across different metrics in voice generative tasks compared to baselines, while handling Russian prosody competitively. These results demonstrate that the derived data notably improves the naturalness of the trained models.

\begin{table*}[ht!]
\footnotesize
    \centering
    \scriptsize
    \caption{Datasets comparison. The \textit{Hours} column indicates the total dataset duration. The \textit{Speech} column denotes the speech type: S (spontaneous), D (dictated), B (books), or Mixed. The \textit{Annotation} column describes the annotation methods: sub (subtitles), man (manual), scr (script), pun (punctuation), mixed text (diverse sources), ASR (automatic speech recognition), str (stresses), ph (IPA phonemes), and ts (word-level timestamps). The highest and second-highest scores are shown in \textbf{bold} and \underline{underlined} text, respectively.}\label{tab:datasets_comparison}
    \begin{tabular}{ c c c c c c c c c c c c }
        \toprule
         Dataset &Hours& Speech & Annotation  & NOI & COL & DIS & LOU & NMOS & UTMOS & MOS $\pm$ 95\% CI & TMR $\pm$ 95\% CI \\ 
        \midrule
        DeepSpeech~\cite{fedoseev2017deepspeech} &6000&S&sub&3.273&3.949&3.4&3.848&3.397&2.483&2.953$\pm$ 0.078&0.701 $\pm$0.084\\
        GOLOS-C~\cite{karpov21_interspeech} &1095&D&man&2.889&3.86&3.237&3.313&3.043&2.393&2.713 $\pm$ 0.077&0.76 $\pm$ 0.069\\
        GOLOS-F~\cite{karpov21_interspeech} &132&D&man&3.293&3.589&2.48&1.597&1.447&1.852&1.477 $\pm$0.142&0.733 $\pm$ 0.072\\
        M-AILABS~\cite{celeste2019mailabs} &46.8 &B&scr+ pun&\underline{4.149}&3.818&\underline{4.012}&3.989&\underline{3.966}&2.765&3.967 $\pm$0.114&\textbf{0.767$\pm$ 0.077}\\
        OpenSTT~\cite{slizhikova2019openstt} &20108&mixed&mixed text&2.588&3.661&2.928&3.191&2.723&2.033&2.633 $\pm$ 0.127&0.759 $\pm$ 0.061\\
        RuLS~\cite{ruls_slr96} &98&B&scr + pun&3.93&4.066&3.757&4.15&3.892&2.8162&3.8 $\pm$ 0.108&0.71 $\pm$ 0.0737\\
        RUSLAN~\cite{10.1007/978-3-030-26061-3_12} &31&B&scr+pun&3.827&\underline{4.266}&3.698&\underline{4.168}&3.788&\underline{2.934}&\underline{4.049$\pm$0.041}&0.766 $\pm$0.063\\
        MCV~\cite{ardila-etal-2020-common} &286&D&scr+pun& 3.511 & 3.859 & 3.391 & 3.755& 3.37 & 2.4456 &{3.875 $\pm$ 0.125}  &  {0.717 $\pm$ 0.072} \\
        SOVA AB~\cite{sova2022dataset} &298&B&scr&3.193&3.694&3.212&3.511&3.096&2.215&2.8 $\pm$0.073&0.74 $\pm$ 0.071\\
        SOVA YT~\cite{sova2022dataset} &17451&S&sub&2.888&3.453&2.854&3.167&2.632&2.567&2.874$\pm$0.095&0.705 $\pm$ 0.082\\
        SOVA D~\cite{sova2022dataset} &191&D&scr&2.859&3.846&3.15&3.442&3.054&2.1883&2.803 $\pm$ 0.086&0.76 $\pm$ 0.069\\
        \textbf{Balalaika (ours)}& 5078&mixed&ASR+pun+str+ph+ts&\textbf{4.235}&\textbf{4.54}&\textbf{4.132}&\textbf{4.391}&\textbf{4.455}&\textbf{3.019}&\textbf{4.601 $\pm$ 0.09}&\textbf{0.767 $\pm$ 0.074}\\
        \bottomrule
    \end{tabular}
\end{table*}

\section{Algorithmic Framework}\label{sec:framework}

\subsection{Audio segmentation}\label{subsec:semantic_vad}

The initial corpus consists of raw audio signals that are unsuitable for direct modeling due to their excessive length. Since arbitrary splitting can disrupt semantic and phonetic continuity, we employ SmartTurnV3.1~\cite{smartturnv2}, a semantic Voice Activity Detection (VAD) model, to identify complete utterances based on context. We filter the resulting segments to ensure data quality, removing any utterances where the speech share is less than 70\% or where internal silence exceeds 1 s. Finally, to optimize input constraints for downstream tasks, we group consecutive segments $s_k$ into chunks capped at 15 seconds, targeting a balanced duration between 5 and 15 seconds per chunk.

\subsection{Audio filtering}\label{subsec:nisqa_pyannotate}

Following segmentation, we apply automatic quality and speaker-purity screening to discard noisy, clipped, and overlapping speech, and to retain only clean single-speaker material for training and evaluation. We first remove utterances shorter than 3 s, then reject segments with excessive impulsive energy by filtering out those whose CREST-factor exceeds a threshold 10~\cite{ITU-T-P501, Chasin2014}. Next, we predict perceptual quality with the NISQA-S model~\cite{deepvk2024nisqa} configured for MOS estimation and discard all segments with MOS (denoted as $q_k$ in tab.~\ref{tab:ablation}) $<$ 4.2. Finally, we use the pyannote speaker diarization pipeline\footnote{\url{https://huggingface.co/pyannote/speaker-diarization-3.1}}~\cite{Bredin23, Plaquet23} to remove any segment containing overlapped speech or two or more speakers, ensuring that each retained utterance contains exactly one active speaker.

\subsection{Transcription}\label{subsec:asrs}

This stage maps each segment   to text by ensembling heterogeneous ASR models and decoding strategies to reduce model-specific errors. For each segment, we generate five hypotheses: (i) \texttt{GigaAM-CTC-v3}~\cite{salute2024gigaam} with CTC decoding, (ii) the same \texttt{GigaAM-CTC-v3} decoded with an external n-gram language model\footnote{\url{https://alphacephei.com/vosk/models/vosk-model-ru-0.22-compile.zip}}~\cite{vosk} to strengthen lexical priors, also producing word-level timestamps, (iii) \texttt{GigaAM-RNNT-v3}~\cite{salute2024gigaam}, (iv) the Russian \texttt{Vosk} model~\cite{vosk}, and (v) \texttt{T-one}~\cite{ttech_tone_2025}.

The final transcript   is obtained using recognizer output voting (ROVER)~\cite{659110}, which aligns all hypotheses into a confusion network and selects a consensus word sequence to minimize word errors. This design combines complementary strengths of CTC vs. RNN-T search/alignment, LM-augmented decoding, and independent model diversity (Vosk and T-one), yielding a more stable transcript for downstream processing.

\subsection{Timestamps}\label{subsec:ctc_timestamps}

We provide word-level timestamps using the CTC+LM decoding output, which yields word-aligned time boundaries. For each segment, we store the ROVER consensus transcript as the primary text annotation, and additionally keep the CTC+LM hypothesis together with its word-level timestamps. This allows downstream applications to use timestamps when needed, while retaining a single, stable transcript for text-based training and evaluation.

\subsection{Punctuation}\label{subsec:rupunct}

We enrich each ASR-consensus transcript with punctuation to encode prosodic structure, which is important for natural Russian TTS intonation. Using RuPunctBig~\cite{RUPunct}, we restore sentence-final and intra-sentential marks directly from the unpunctuated word sequence produced upstream. In our pipeline, explicit punctuation serves as a lightweight proxy for phrase breaks and discourse structure, and is consistently associated with improved intonational naturalness in synthesized speech (Sec.~\ref{sec:res_rq3}).

\subsection{Stress placement and \textipa{\H{e}}-normalization}\label{subsec:ruaccent}

We enrich each punctuated transcript with lexical stress marks and resolve the Russian \textipa{e}/\textipa{\H{e}} ambiguity to support prosody modeling and homograph disambiguation (Sec.~\ref{sec:res_rq3}). We apply RuAccent~\cite{petrov-2025-ruaccent} for context-aware stress assignment and to restore \textipa{\H{e}} when required by pronunciation. This yields a pronunciation-consistent text representation that improves downstream synthesis stability and naturalness.

\subsection{Phonemization}\label{subsec:g2p}

This stage converts stress-normalized text into IPA phoneme sequences while preserving punctuation and stress marks. We apply a word-level G2P model that transliterates graphemes to phonemes and copies non-graphemic symbols unchanged.

Our G2P is a lightweight transformer encoder–decoder~\cite{Yolchuyeva_2019} trained with a BPE tokenizer on graphemes and an IPA target inventory derived from Wiktextract pronunciations~\cite{ylonen-2022-wiktextract}. The model uses \texttt{d\_model} = 128, \texttt{d\_ff} = 512, 3 encoder and 3 decoder layers, 4 attention heads, and a 64-token limit; we train with AdamW (lr $3\times10^{-4}$), batch size 256, label smoothing 0.1 for 10 epochs, and decode greedily at inference. Using IPA targets improves coverage of Russian phonology (e.g., vowel reduction and consonant devoicing), yielding stable phoneme sequences for downstream modeling.

\subsection{Dataset: Balalaika}

Balalaika is assembled from multiple publicly available Russian speech and text resources; we provide the full provenance and per-source breakdown (including Yandex Music, OpenSTT~\cite{slizhikova2019openstt}, ESpeech~\cite{Den4ikAI_ESpeechTechReport_2025}, GOLOS~\cite{karpov21_interspeech}, DeepSpeech~\cite{fedoseev2017deepspeech}, and Biggest Russian Books~\cite{its5q_biggest_ru_book}). The data was processed by our annotation framework, resulting in complete multi-layer annotations for all retained segments. The total size of the dataset is approximately 5078  hours of audio.

We provide all necessary scripts to reproduce the dataset’s annotations from the original sources, (for Yandex Music split)\footnote{\url{https://hf.co/datasets/lab260/Balalaika2000H}}, and we release the complete end-to-end annotation pipeline\footnote{\url{https://github.com/lab260ru/balalaika}}. Original audio remains distributed under the respective source licenses.
\section{Experimental Setup}\label{sec:experimental_setup}

\subsection{MOS Evaluation}

\begin{table*}[ht!]
\footnotesize
   \scriptsize
    \centering
    \caption{ {Comparison of speech synthesis models trained on different datasets. The highest and second-highest scores for the metrics are shown in \textbf{bold} and \underline{underlined} text.}}\label{tab:tts}
    \begin{threeparttable}
    \begin{tabular}{c c c c c c c c c c c }
         \toprule
         Dataset  & NOI & COL & DIS & LOU & NMOS& TTS MOS & UTMOS & MOS $\pm$ 95\% CI & IntMOS $\pm$ 95\% CI & CER \\
         \midrule
         DeepSpeech~\cite{fedoseev2017deepspeech} & 2.6946 & 3.6406 & 2.6819 & 3.5315 & 2.7796 & 2.4468 &1.4871 & {1.597 $\pm$ 0.078}& {0.62 $\pm$ 0.098}& 0.7693 \\
         
         GOLOS-C~\cite{karpov21_interspeech} & 2.2051 & 3.6959 & 2.1849 & 1.9105 & 1.7415 &1.9672& 0.9203 & {0.615 $\pm$ 0.064}& {0.034 $\pm$ 0.04}& 0.9999 \\
         
         GOLOS-F~\cite{karpov21_interspeech} & 3.4367 & 3.6482 & 2.4353 & 1.4922 & 1.1814 & 1.6324 & 0.8445 & {0.036 $\pm$ 0.024}& {0.013 $\pm$ 0.026}& 1 \\
         
         M\_AILABS~\cite{celeste2019mailabs} & 3.7924 & 3.3723 & 3.6128 & 3.8226 & 3.5301  & 3.0321 & 2.30653 & {2.962 $\pm$ 0.052}& {2.208 $\pm$ 0.071}& \underline{0.0908} \\
         
         RUSLAN~\cite{10.1007/978-3-030-26061-3_12} & \underline{4.0193} & 4.3278 & 3.7862 & \underline{4.2483} & 3.7438  & 1.9134 & 2.12968 & {3.253 $\pm$ 0.068}& {\textbf{3.182 $\pm$ 0.094}}& \textbf{0.0496}  \\
         
         OpenSTT~\cite{slizhikova2019openstt} & 2.1172 & 3.7438 & 2.5426 & 3.1770 & 2.4066 & 1.6217 & 1.6762 &  {1.259 $\pm$ 0.058}& {0.135 $\pm$ 0.053}& 0.96 \\
         
         RuLS~\cite{ruls_slr96} & 3.9336 & 4.2351 & 3.5956 & 3.9382 & \underline{3.7876}  & 2.9269 & 2.1068 & {2.75 $\pm$ 0.094}& {2.142 $\pm$ 0.086}& 0.1003 \\
         
         MCV~\cite{ardila-etal-2020-common} & 4.0167 & 4.2227 & 3.6504 & 3.9415 & 3.7557 & 3.2095 & 2.12294 & {2.749 $\pm$ 0.062}& {2.462 $\pm$ 0.094}& 0.238  \\
         
         SOVA AB~\cite{sova2022dataset} & 2.9993 & 4.1601 & 2.9049 & 3.5548 & 2.9691 & 2.5572 & 1.49 & {1.354 $\pm$ 0.063}& {0.156 $\pm$ 0.049}& 0.9112  \\
         
         SOVA YT~\cite{sova2022dataset} & 2.0664 & 3.0234 & 1.5248 & 2.5048 & 1.4161 & 1.5147 & 0.8036 & {0.979 $\pm$ 0.016}& {0.018 $\pm$ 0.017}& 0.9998\\
         
         SOVA D~\cite{sova2022dataset} & 2.3487 & 4.0085 & 2.8124 & 3.4032 & 2.7414 & 2.2699 & 1.43 & {1.336 $\pm$ 0.039}& {0.196 $\pm$ 0.062}& 0.8714  \\
         
          \textbf{Balalaika (ours)}  & \textbf{4.2002}  & \textbf{4.5799} &  \textbf{4.3504} & \textbf{4.4632} & \textbf{4.4843} & \textbf{3.5703} & \textbf{2.73848} & {\textbf{3.618 $\pm$ 0.083}}& {\underline{2.532 $\pm$ 0.09}}& 0.1062  \\
         
         \bottomrule
    \end{tabular}
    \end{threeparttable}
\end{table*}
\begin{table*}[ht]
\tiny
    \centering
    \caption{ {Ablation study of different configurations of our dataset. The highest and second-highest scores for the metrics are shown in \textbf{bold} and \underline{underlined} text.}}\label{tab:ablation}
    \begin{threeparttable}
    \begin{tabular}{c c c c c c c c c c c }
        \toprule
        Dataset & NOI & COL & DIS & LOU & NMOS& TTS MOS & UTMOS & MOS $\pm$ 95\% CI & IntMOS $\pm$ 95\% CI & CER  \\
        \midrule
        ours (MOS$>4.2$) &  4.0555 &  4.5158 & 4.2646 &  4.3337 &  4.3456 & 3.3155 & 2.64739 & {3.41 $\pm$ 0.081}& {2.305 $\pm$ 0.088}& 0.1347\\
        
        \begin{tabular}{@{}c@{}}ours + stresses ($q_k>4.2$)\end{tabular} & 4.1358 &  4.5275 &  4.2979 &  \underline{4.4175} &  \underline{4.4446} & 3.5472  & \underline{2.73128} & \underline{3.522 $\pm$ 0.082}& \underline{2.48 $\pm$ 0.094}&  0.1291  \\
        
        \begin{tabular}{@{}c@{}}ours +  punctuation ($q_k>4.2$)\end{tabular} & \underline{4.1533}  & \underline{4.547} & \underline{4.3249} & 4.4148 & 4.4326 & \underline{3.5671}  & 2.63233 &  { 3.44 $\pm$ 0.079}&  {2.448 $\pm$ 0.089}& \underline{0.1123}  \\

        ours +  stresses +  punctuation ($q_k>3$) & 2.4276 & 4.2440 & 3.8387 & 3.5028 & 2.6816 & 2.708 & 2.17835 & {2.446 $\pm$ 0.079}& {1.877 $\pm$ 0.094}& 0.2704  \\
         
         ours +  stresses +  punctuation ($q_k>3.5$) & 3.3053 & {4.3606} & {4.0656} & 3.9652 & 3.6308 & {3.5002} &  {2.48488} & {3.256 $\pm$0.095}& {2.377 $\pm$ 0.095}& 0.171 \\

         \begin{tabular}{@{}c@{}}ours +  stresses +  punctuation ($q_k>4.2$)\end{tabular}  &  \textbf{4.2002}  & \textbf{4.5799} &  \textbf{4.3504} & \textbf{4.4632} & \textbf{4.4843} & \textbf{3.5703} & \textbf{2.73848} & \textbf{3.618 $\pm$ 0.083} & \textbf{2.532 $\pm$ 0.09}& \textbf{0.1062}   \\
        \bottomrule
    \end{tabular}
    \end{threeparttable}
\end{table*}

We collected human ratings using LabelSpeech\footnote{\url{https://github.com/lab260ru/LabelSpeech}}~\cite{LabelSpeech}. Raters assigned MOS on a 0--5 scale (0: non-speech; 1: low-quality telephony; 2: very low quality; 3: unprofessional recording; 4: studio quality with artifacts; 5: perfect studio quality).

\begin{table}[ht!]
\footnotesize
    \scriptsize
    \centering
    \caption{ {Comparison of speech denoising models trained on different datasets. The highest and second-highest scores for the metrics are shown in \textbf{bold} and \underline{underlined} text.}}    \label{tab:restoration}
    \begin{threeparttable}
    \begin{tabular}{p{1.7cm} p{0.3cm} p{0.3cm} p{0.3cm} p{0.3cm} p{0.3cm} p{0.3cm} p{0.3cm} p{0.5cm}}
         \toprule
         Dataset & CSIG & CBAK & COVL & PESQ & VISQ & UTM & STOI & SISDR \\
         \midrule
         test subset & 2.926 & 2.506 & 2.127 & 1.456 & 3.523 & 2.250 & 0.877 & 7.582  \\
         \midrule 
         DeepSpeech~\cite{fedoseev2017deepspeech} & 3.678 & 2.800 & 3.154 & 2.541 & 3.957 & 2.406 & 0.923 & \underline{8.577}\\
         GOLOS-C~\cite{karpov21_interspeech} & 3.715 & 2.684 & 3.190 & 2.504 & 3.974 & 2.409 & 0.922 & 8.374 \\
         GOLOS-F~\cite{karpov21_interspeech} & 3.760 & 2.663 & 3.170 & 2.350 & 3.946 & 2.380 & 0.917 & 7.786 \\
         M\_AILABS~\cite{celeste2019mailabs} & 3.778 & 3.084 & \underline{3.287} & \underline{2.721} & 4.012 & 2.594 & \underline{0.930} & 8.257 \\
         RUSLAN~\cite{10.1007/978-3-030-26061-3_12} & 3.530 & 2.893 & 3.024 & 2.476 & 3.804 & \textbf{2.621} & 0.878 & 5.833 \\
         OpenSTT~\cite{slizhikova2019openstt} & 3.688 & 2.953 & 3.195 & 2.634 & 3.931 & 2.425 & 0.924 & 8.051 \\
         RuLS~\cite{ruls_slr96} & 3.740 & 2.912 & 3.204 & 2.584 & 3.981 & 2.537 & 0.925 & 8.717 \\
         MCV~\cite{ardila-etal-2020-common} & \underline{3.785} & \underline{3.116} & 3.228 &  2.651 & 4.005 & 2.520 & 0.926 & 8.113 \\
         SOVA AB~\cite{sova2022dataset} & 3.780 & 3.088 & 3.258 & 2.644 & \underline{4.032} & 2.420 & 0.928 & 8.744 \\
         SOVA YT~\cite{sova2022dataset} & 3.678 & 2.970 & 3.188 & 2.610 & 3.974 & 2.414 & 0.926 & 8.170 \\
         SOVA D~\cite{sova2022dataset} & 3.743 & 2.854 & 3.227 & 2.582 & 3.987 & 2.471 & 0.924 & 8.317 \\
         \textbf{Balalaika (ours)} & \textbf{3.856} & \textbf{3.165} & \textbf{3.340} & \textbf{2.723} & \textbf{4.036} & \underline{2.614} & \textbf{0.931} & \textbf{8.809} \\
         \bottomrule
    \end{tabular}
    \end{threeparttable}
\end{table}

Since we posit that additional annotations affect the quality of synthetic speech, we additionally collected an intonation-focused MOS (IntMOS) to capture prosodic naturalness beyond overall quality. Raters assigned MOS on a 0--5 scale (0: unintelligible; 1: clearly unnatural; 2: mostly robotic; 3: indeterminate; 4: dictation-like; 5: natural conversational speech).

Each audio item in our human-feedback evaluations received at least seven independent ratings from native speakers of Russian. For each clip, we aggregated ratings by taking the median. System-level scores were computed as the mean across clips, and we report 95\% confidence intervals (CI). IntMOS is reported alongside MOS to isolate prosodic and intonational aspects that are specifically targeted by our annotation layers, providing a direct test of their contribution to perceived naturalness.

\subsection{RQ1: Framework Comparison}

We evaluate RQ1 by comparing datasets. For objective assessment, we apply the original NISQA model~\cite{Mittag_Naderi_Chehadi_Möller_2021}, reporting its five metrics on each dataset: noisiness (NOI), coloration (COL), discontinuity (DIS), loudness (LOU), and overall MOS (NMOS). We additionally report the UTokyo-SaruLab MOS Prediction System(UTMOS)~\cite{baba2024utmosv2}. For human evaluation, we use classical MOS. We also quantify transcript fidelity via a manual Text Match Rate (TMR), defined as the percentage of clips whose displayed text matches what is spoken according to rater judgment. To keep comparisons balanced, we draw 200 items per dataset for MOS and TMR, while NISQA and UTMOSv2 are computed over the full available audio. We compare against 11 public Russian corpora~\cite{fedoseev2017deepspeech, karpov21_interspeech, celeste2019mailabs, slizhikova2019openstt, ruls_slr96, 10.1007/978-3-030-26061-3_12, sova2022dataset, ardila-etal-2020-common}.


\subsection{RQ2: Speech Denoising Performance}

To isolate the effect of data quality, we train SEMamba~\cite{chao2024investigation} from scratch on each dataset under an identical data/training budget. 
All runs use the same recipe (Adam, lr=$5{\times}10^{-4}$, batch=8, 50k steps) and are evaluated at the final checkpoint (no early stopping). 
Training uses 25\,h randomly subsampled per dataset (disjoint from test) with MUSAN noise~\cite{musan2015} and RIR augmentation~\cite{7953152} applied uniformly.

The test benchmark contains 3,000 samples: 500 clips each from M-AILABS, RUSLAN, RuLS, and from our dataset, drawn from their held-out test splits to ensure no leakage. In accordance with the methodology of the original paper~\cite{chao2024investigation}, the following objective metrics were selected for evaluation: prediction of the signal distortion(CSIG)~\cite{hu06_interspeech}, prediction of the background intrusiveness(CBAK)~\cite{hu06_interspeech}, prediction of the overall speech quality(COVL)~\cite{hu06_interspeech}, perceptual evaluation of speech quality(PESQ)~\cite{941023}, short-time objective intelligibility measure(STOI)~\cite{5713237}. Along with this, we used virtual speech quality objective listener(VISQOL)~\cite{9123150} to evaluate audio quality. To evaluate the signal distortion ratio, we used scale invariant signal\textbf{d} distortion ratio(SI-SDR).

\subsection{RQ3: Text-to-Speech performance}\label{subsec:exp_setup_rq5}

To isolate the effect of training data, we train a single VITS model~\cite{pmlr-v139-kim21f} on each dataset under an equalized data and training budget: 25\,h of audio and a fixed recipe (Adam, lr=$10^{-4}$, batch=32, 100k steps), evaluated at the final checkpoint (no early stopping).

Evaluation uses a heterogeneous held-out set of 2,000 texts from Balalaika’s multi-source test subsets, disjoint from the sampled training subsets and shared across systems. Objective quality is assessed with NISQA and NISQA-TTS (TTS MOS)~\cite{mittag20_interspeech}; we also report UTMOS. Intelligibility is measured by Character Error Rate (CER), where predicted text computed by transcribing synthetic audio with GigaAMv3-RNNT~\cite{salute2024gigaam}. Subjective quality is measured with classical MOS and IntMOS on a randomly sampled 200-text subset.

We analyze how individual annotation layers affect TTS quality by training models on the same audio while varying only the text annotations. Using the pipeline and evaluation protocol from Sec.~\ref{subsec:exp_setup_rq5}, we compare six settings - plain transcripts, +stress, +punctuation, +stress+punctuation, and two different MOS filtering thresholds - to isolate the impact of each annotation choice on naturalness and intelligibility.











\section{Results and Discussion}\label{sec:results}

\subsection{RQ1: Framework Comparison}

Our annotation framework produces higher-quality Russian speech datasets than existing methods in perceptual quality and text–audio consistency, Table \ref{tab:datasets_comparison}. The unified Balalaika dataset attains the best scores across all objective predictors and human MOS and IntMOS, outperforming public corpora such as M-AILABS~\cite{celeste2019mailabs}, RuLS~\cite{ruls_slr96}, and RUSLAN~\cite{10.1007/978-3-030-26061-3_12} on every reported metric; manual TMR likewise ranks it at the top, supporting the reliability of the multi‑ASR fusion and confirming that the pipeline yields superior audio quality and transcript fidelity through effective annotation.

\subsection{RQ2: Speech Denoising Performance}

Training denoisers from scratch on our data yields superior performance to models trained on alternative Russian datasets, with the SEMamba~\cite{chao2024investigation} trained on our dataset achieving the top score across most objective metrics in Table \ref{tab:restoration}, indicating simultaneous gains in perceived signal quality, background suppression, overall quality, and intelligibility compared to systems trained on M-AILABS~\cite{celeste2019mailabs}, RuLS~\cite{ruls_slr96}, MCV~\cite{ardila-etal-2020-common}, OpenSTT~\cite{slizhikova2019openstt}, and SOVA variants~\cite{sova2022dataset}.

These gains are mirrored in SI‑SDR, where the model trained on our unified dataset achieves the strongest distortion reduction; models trained on our data remain competitive across metrics and generally surpass counterparts trained on other corpora, underscoring that denoising benefits from the higher‑quality data produced by the framework.

\subsection{RQ3: Text-to-Speech performance}\label{sec:res_rq3}

Training TTS models on Balalaika yields the highest objective quality scores and human MOS in Table~\ref{tab:tts}, while maintaining competitive CER. RUSLAN achieves the highest IntMOS, whereas Balalaika ranks second; several other differences in subjective scores are smaller than the corresponding confidence intervals. These results indicate improved overall naturalness, but do not establish a strict ordering among systems with overlapping intervals.

For intonational naturalness, our system ranks second only to the single-speaker RUSLAN model, consistent with the expressivity advantage of single-speaker data; overall, the combined MOS, objective metrics, and CER yield the strongest naturalness - intelligibility trade-off. Prosody-aware annotations provide additional gains: Table~\ref{tab:ablation} shows that punctuation and stress are most effective jointly, improving MOS/IntMOS and reducing CER compared to either layer alone.

Finally, stricter quality filtering improves synthesis: raising the MOS threshold from 3.5 to 4.2 increases predicted and human-rated naturalness, improves IntMOS, and lowers CER, whereas MOS$>$3 degrades all metrics, suggesting low-quality audio artifacts propagate to both prosody and intelligibility.

\subsection{Limitations}

All systems were trained under the same fixed data and compute budget and were not run to full convergence. This ensures a fair cross-dataset comparison, but some models may remain under-trained and could achieve higher absolute scores with longer training.

The current pipeline is also Russian-specific, relying on language-dependent components (ASR, punctuation restoration, stress placement, and G2P), which limits direct transfer to other languages. However, the pipeline is modular, so extending it to new languages reduces to swapping in comparable tools as they become available.

The TTS test set is a heterogeneous held-out split of Balalaika shared by all systems. Because several source datasets are also represented in Balalaika, models trained on those datasets may benefit from partial domain alignment. The experiment therefore provides a controlled comparison on a common test set, but not a fully source-independent evaluation.

\section{Conclusion}\label{sec:conclusion}

Balalaika combines semantic segmentation, quality and speaker filtering, multi-ASR ROVER transcription, and prosody-aware text enrichment to build a 5.1k-hour Russian corpus. Under equalized budgets, its data improves denoising and TTS, with stress plus punctuation giving the strongest synthesis quality; we release the pipeline and reproducibility scripts.


\clearpage

\section{Generative AI Use Disclosure}
Generative AI tools were used only for language editing and manuscript polishing (e.g., improving clarity, grammar, and style). They were not used to generate substantial technical content, including the proposed methods, experiments, results, figures/tables, or conclusions. All authors reviewed and approved the final manuscript and take full responsibility for the content.
\bibliographystyle{IEEEtran}
\bibliography{mybib}

\ifcameraready
     \clearpage
     \appendix
    \section{Related Work}\label{sec:related_work}

\subsection{Speech Datasets and Annotation}

\begin{table*}[ht!]
\scriptsize
    \centering
    \caption{Datasets comparison}\label{tab:datasets_related_work}   \begin{threeparttable}
    \begin{tabular}{c c c c c c c c c c c}
        \toprule
         Dataset & Hours & Speech Type & Licensing & PUN & STR & Phonemes & Timestamps & Multispeaker & Text annotation &  Scalability \\ 
        \midrule
        LibriTTS-R(EN)\cite{koizumi23_interspeech} & 585 & B & CC BY 4.0 & $\checkmark$ & $\checkmark$ & $\checkmark$ & $\checkmark$ & $\checkmark$ & scr & partial\\
        \makecell{WeNetSpeech4TTS \\ (ZH)\cite{ma24d_interspeech} } & 12800 & S &CC BY 4.0& $\times$ & $\times$ & $\times$ &$\checkmark$ & $\checkmark$ & ASR & yes \\
        TTSops\cite{seki2025ttsopsclosedloopcorpusoptimization}(JP) & 66 & S & not open & $\checkmark$& $\times$ & $\times$ & $\checkmark$ & $\checkmark$ & ASR, sub & yes  \\
        \midrule
        DeepSpeech~\cite{fedoseev2017deepspeech} & 6000 & S & MPL 2.0 & $\times$ & $\times$ & $\times$ & $\times$ & $\checkmark$ & sub & partial\\ 
        GOLOS-C~\cite{karpov21_interspeech} & 1095 &D & Golos$^a$ & $\times$ & $\times$ & $\times$ & $\times$& $\checkmark$ & man & no  \\ 
        GOLOS-F~\cite{karpov21_interspeech} & 132 & D & Golos$^a$ & $\times$ & $\times$ & $\times$ & $\times$& $\checkmark$ & man & no  \\
        M-AILABS~\cite{celeste2019mailabs} & 46.8 & B & BSD-3-Clause & $\checkmark$ & $\times$ & $\times$ & $\times$ & $\checkmark$ & scr & no\\ 
        OpenSTT~\cite{slizhikova2019openstt} & 20108 & S, B, SY, D & CC-BY-NC & $\times$ & $\times$ & $\times$ & $\times$ & $\checkmark$ & \makecell{sub, ASR, \\ scr, TTS} & partial \\ 
        RuLS~\cite{ruls_slr96} & 98 & B & PD$^b$ & $\checkmark$ & $\times$ & $\times$ & $\times$ & $\checkmark$ & scr & no\\ 
        RUSLAN~\cite{10.1007/978-3-030-26061-3_12} & 31 & D & \makecell{CC BY-NC-SA \\ 4.0} &  $\checkmark $ & $\times$ & $\times$ & $\times$ & $\times$ & scr & no \\ 
        MCV~\cite{ardila-etal-2020-common} & 286 & D & CC0 & $\checkmark$ & $\times$ & $\times$ & $\times$ & $\checkmark$ & scr & partial\\ 
        SOVA AB~\cite{sova2022dataset} & 298 & B & CC BY 4.0 & $\times$ &$\times$ & $\times$ & $\times$ & $\checkmark$ & scr & no\\ 
        SOVA YT~\cite{sova2022dataset} & 17451 & S & CC BY 4.0 & $\times$ & $\times$ & $\times$ & $\times$ & $\checkmark$ & sub & partial\\ 
        SOVA D~\cite{sova2022dataset} & 191 & D & CC BY 4.0 & $\times$ & $\times$ & $\times$ & $\times$ & $\checkmark$ & scr & no\\ 
        \midrule
        \textbf{Ours} & 5078 & S, B, D & mixed$^c$ & $\checkmark$ & $\checkmark$ & $\checkmark$ & $\checkmark$ & $\checkmark$ & ASR & yes\\ 
        \bottomrule
    \end{tabular}
    \begin{tablenotes}
         \item \textbf{Note:} This table compares speech datasets for TTS training, contrasting high-resource languages (EN: English, ZH: Chinese, JP: Japanese) with Russian ones on scale, annotations, and scalability for web audio mining. Attributes include Hours (dataset size), Speech Type (B = audiobook; S = spoken; D = dictated; SY = synthesized), Licensing, and features ($\checkmark$ = present, $\times$ = absent), including PUN (punctuation annotations) and STR (stress annotations). Text Annotation: scr = scripted (dictated from text); ASR = via Automatic Speech Recognition; sub = subtitles from video hosting; TTS = synthesized from text; man = manual. Scalability: yes = fully automated; partial = limited (e.g., manual intervention); no = non-scalable.
        \item $^a$ Custom SberDevices license (similar to CC BY-SA). Full details available at \url{https://github.com/sberdevices/golos/blob/master/license/en_us.pdf}.
        \item $^b$ The dataset is Public Domain in the USA. 
        \item $^c$ The code is released under the CC BY-NC-SA 4.0 license, the annotations under CC BY-NC-ND 4.0, and the original audio distributed under its respective licenses.
    \end{tablenotes}
    \label{tab:datasets_comparison}
    \end{threeparttable}
\end{table*}

The rapid growth of web audio~\cite{EdisonResearch2025, SNSInsider2025} has made speech data central to multilingual voice interfaces and generative applications~\cite{10.1145/3704262}. Yet many speech datasets lack scalability and rich annotations, especially for under-resourced languages~\cite{geng-etal-2025-scaling}, where simplistic curation misses prosody and phonetic detail critical for TTS~\cite{lau2025dataqualityissuesmultilingual}. For Russian, these issues are compounded by vowel reduction~\cite{RUDN:TLSS34176}, mobile stress~\cite{petrov-2025-ruaccent}, and intonation patterns absent in dictated or synthesized speech~\cite{zhang23_ssw}, leading to biases that hinder web-scale use and inclusivity~\cite{zee-etal-2024-group, FENG2024101567}. As Table \ref{tab:datasets_related_work} shows, audiobook and dictated corpora (e.g., M-AILABS~\cite{celeste2019mailabs}, RUSLAN~\cite{10.1007/978-3-030-26061-3_12}) provide scripted text but omit stress and phonemes and often sound unnatural for spontaneous speech, while RuLS~\cite{ruls_slr96} and GOLOS~\cite{karpov21_interspeech} emphasize manual or scripted transcription with far-field noise, limiting prosody-aware mining despite openness.

Spontaneous and crowd-sourced corpora such as OpenSTT~\cite{slizhikova2019openstt} and SOVA subsets~\cite{sova2022dataset} rely on ASR or subtitles but omit prosodic cues, and often suffer from background noise. Most Russian datasets lack stress, phonemes, and timestamps, and many also lack punctuation, with scalability frequently “partial” or “no” due to manual pipelines. Subtitle-dependent sets (e.g., DeepSpeech~\cite{fedoseev2017deepspeech}, SOVA YT~\cite{sova2022dataset}) require pre-existing text, MCV’s~\cite{ardila2020commonvoicemassivelymultilingualspeech} manual workflows limit flexibility, and OpenSTT’s~\cite{slizhikova2019openstt} parsing focus constrains spontaneous sources - together overlooking prosody and spontaneous speech and compounding web-scale mining challenges.

High-resource corpora illustrate stronger annotation pipelines: WenetSpeech4TTS~\cite{ma24d_interspeech} offers 12800 hours of spontaneous Mandarin with ASR, denoising, and word-level timestamps but no explicit stress/phonemes; LibriTTS-R~\cite{koizumi23_interspeech} provides 585 hours of English audiobook speech with phoneme alignments, timestamps, and G2P-inferred stress, yet its approach does not transfer to Russian due to homographs and movable stress; and TTSOps~\cite{seki2025ttsopsclosedloopcorpusoptimization} yields ~66h of Japanese spontaneous speech from noisy YouTube via adaptive cleaning, subtitle-based ASR, and evaluation-in-the-loop, achieving high naturalness while highlighting gaps in annotation depth and scalability for Russian.  These gaps in Russian speech datasets highlight the need for data-centric frameworks. 

\subsection{Annotation Pipelines and Frameworks in Speech Processing}

Speech annotation pipelines play a pivotal role in web audio processing, enabling scalable mining of vast online resources like podcasts. General frameworks~\cite{ma24d_interspeech, he2025emilialargescaleextensivemultilingual, song2024touchttsembarrassinglysimpletts, seki2025ttsopsclosedloopcorpusoptimization} often incorporate modular components such as ASR for transcription, voice activity detection (VAD) for segmentation, and quality filtering metrics to ensure data usability.

In contrast, Russian-specific annotation pipelines typically rely on manual curation~\cite{karpov21_interspeech, ardila2020commonvoicemassivelymultilingualspeech} or simplistic methods like subtitle alignment~\cite{fedoseev2017deepspeech,slizhikova2019openstt} and ASR for transcription~\cite{slizhikova2019openstt}, often lacking prosody-aware features and scalability. To our knowledge, our modular framework is the first of its kind for Russian, integrating prosody-focused annotation with scalable components for data mining.

Unlike previous approaches for Russian that stop at bare text-audio pairs, our framework adds punctuation, lexical stress, and word-level durations, supplying TTS models with rich prosodic cues. A multi-step quality filter removes unusable clips automatically. Each stage -- is exposed as a self-contained component, allowing language-specific modules to be swapped. Together these innovations deliver the first prosody-aware, scalable Russian annotation framework.

    \section{Problem Definition and Notation}\label{sec:problem_definition}

\begin{algorithm}[H]
\caption{Utterance segmentation with speech/silence filtering}
\label{alg:segmentation}
\begin{algorithmic}
\Require $\mathcal{A}=\{a_i\}_{i=1}^{M}$; $d_{\max}$ 
\Ensure $D=\{(s_k)\}_{k=1}^{N}$ 

\State $D \gets \emptyset$ 
\For{$i \gets 1$ \textbf{to} $M$} 
  \State $(\{t^s_j\}, \{t^e_j\}) \gets \texttt{SemanticVAD}(a_i)$ \Comment{Speech regions; $j \in \{1,\ldots,P_i\}$}
  \State pieces $\gets$ [ ] \Comment{Candidate index ranges over VAD regions}
  \State $n \gets P_i$ \Comment{Number of speech regions in $a_i$}
  \State $\text{start\_idx} \gets 0$

  \While{$\text{start\_idx} < n$} 
    \State $L \gets t^s_{\text{start\_idx}}$
    \State $\text{end\_idx} \gets \text{start\_idx}$

    \While{$\text{end\_idx} < n$} \Comment{Extend until exceeding $d_{\max}$}
      \State $R \gets t^e_{\text{end\_idx}}$ 
      \If{$(R - L) > d_{\max}$} \Comment{Stop if too long}
        \State \textbf{break}
      \EndIf
      \State $\text{end\_idx} \gets \text{end\_idx} + 1$ \Comment{Try adding the next region}
    \EndWhile

    \If{$\text{end\_idx} > \text{start\_idx}$} 
      \State $\Delta \gets t^e_{\text{end\_idx}-1} - L$ \Comment{Candidate duration}
      \If{$\Delta \ge d_{\max}/3$ \textbf{ and } $\Delta \le d_{\max}$} \Comment{Keep only reasonable durations}
        \State pieces.append$(\text{start\_idx}, \text{end\_idx}-1)$ 
      \EndIf
    \EndIf

    \State $\text{start\_idx} \gets \max(\text{end\_idx}, \text{start\_idx}+1)$ \Comment{Advance pointer; avoid stalling}
  \EndWhile

  \ForAll{$(u,v)$ \textbf{in} pieces} \Comment{Validate each candidate}
    \State $\ell \gets t^e_v - t^s_u$ \Comment{Total segment length}
    \State $\rho \gets \frac{\sum_{j=u}^{v}(t^e_j - t^s_j)}{\ell}$ \Comment{Speech share in the segment}
    \State $\sigma \gets \max_{j\in\{u,\ldots,v-1\}}(t^s_{j+1}-t^e_j)$ \Comment{Max internal silence gap}
    \If{$\rho \ge 0.7$ \textbf{ and } $\sigma \le 1\text{s}$} \Comment{Accept only mostly-speech segments without long pauses}
      \State $D \gets D \cup \{a_i[t^s_{u}:t^e_{v}]\}$ \Comment{Extract waveform slice and add to output}
    \EndIf
  \EndFor
\EndFor

\State \Return $D$ \Comment{Return accepted segments}
\end{algorithmic}
\end{algorithm}

Our primary objective is to create a scalable, fully automated pipeline that processes Russian speech from diverse sources, filters it for quality, and enriches it with essential annotations, transforming raw audio into structured datasets. Focusing on Russian-specific phonetics, the workflow delivers high-fidelity data at scale with no manual intervention.
\begin{algorithm}[H]
\caption{Whole pipeline}
\label{alg:pipeline}
\begin{algorithmic}
\Require  $\mathcal{A}=\{a_i\}_{i=1}^{M}$;  $d_{\max}; \theta_q; v_d$
\Ensure  $\mathcal{D} = \{(s_k, \mathbf{T}_k, \mathbf{A}_k\}_{k=1}^N$
\State $\mathcal{D} \gets \emptyset$
\State $D \gets \mathrm{Segmentation}(\mathcal{A}, d_{\max})$ \Comment{alg. \ref{alg:segmentation}}

\For{each $s_k \in D$} 
  \State $q_k = \texttt{QualityEstimator}(s_k)$ 
  \State $d_k = \texttt{SingleSpeakerDetector}(s_k)$ 
  \If{$q_k \ge \theta_q$ \textbf{ and } $d_k = v_d$}
      \State $r_k^{CTC} = ASR_{CTC}(s_k)$ 
      \State $r_k^{CTC+LM}, \tau_k=ASR_{CTC+LM}(s_k)$ 
      \State $r_k^{RNNT}=ASR_{RNNT}(s_k)$ 
      \State $r_k^{VOSK}=ASR_{VOSK}(s_k)$ 
      \State $r_k^{TONE} = ASR_{TONE}(s_k)$
      \State $ T_k^r = \texttt{ROVER}({r_k^{CTC}, r_k^{CTC+LM}, r_k^{RNNT}, r_k^{Vosk}, r_k^{TONE}})$ 
      \State $I_k= \textbf{1}_k(T_k^r, r_k^{CTC})$ 
      \State $c_k = CREST(s_k)$
      \If{$I_k=1$ and $c_k<10$ and $l_k>3$}
          \State $T_k^p = \texttt{Punctuator}(T_k^r)$ 
          \State $T_k^s = \texttt{StressPlacer}(T_k^p)$ 
          \State $\Phi_k = \texttt{G2P}(T_k^s)$ 
          \State $\mathbf{T}_k \gets \{T_k^r, T_k^p, T_k^s, \Phi_k\}$; 
          \State $\mathbf{A}_k \gets \{\tau_k, q_k, d_k\}$
        \State $\mathcal{D} \gets \mathcal{D} \cup \{(s_k, \mathbf{T}_k, \mathbf{A}_k)\}$
      \EndIf
  \EndIf
\EndFor

\State \Return $\mathcal{D}$
\end{algorithmic}
\end{algorithm}

Let $\mathcal{A} = \{ a_i\}^M_{i=1}$ denote the raw web audio corpus, where each $a_i$ is an unprocessed audio signal from diverse sources with variable lengths and no initial annotations. The pipeline transforms each $a_i$ into a structured dataset $\mathcal{D} = \{s_k, \mathbf{T}_k, \mathbf{A}_k\}^N_{k=1}$, where $s_k$ is a segmented audio clip; $\mathbf{T}_k = \{T_k^r, T_k^t, T_k^p, T_k^s, \Phi_k\}$ includes textual annotations, with $T_k^r$ as the raw transcript,  $T_k^t$  as the trancript from $ASR^{CTC+LM}$, $T_k^p$ as the punctuated text, $T_k^s$ as the stress-annotated text, and $\Phi_k$ as the phoneme sequence; and $\mathbf{A}_k = \{\tau_k, q_k, d_k, c_k, l_k\}$ captures auxiliary annotations, where $\tau_k$ is word-level timestamps, $q_k$ is the quality score, $d_k$ indicates single-speaker status, $c_k$ is the CREST factor and $l_k$ is the total utterance length.

Our evaluation goals assess the framework's outputs in data quality, model performance, and annotation impacts for generative tasks.

We compare our datasets to Russian ones on audio quality and understandability (RQ1). We assess denoising models trained on our data against others (RQ2). For TTS, we measure quality, understandability and intonation (RQ3).  We evaluate speech restoration models trained on our data versus baseline approaches (RQ4). Finally, we measure stage-wise processing throughput (RQ5).

    \section{MOS Evaluation in details}\label{sec:dataset_balalaika}

We collected human ratings using Anonymised Platform\footnote{To ensure authors' anonimity, the platform link well be available at camera-ready version}. Raters received the following rubric for manual MOS:\\
\textbf{"5" --} Perfect studio quality: clear sound without noise, reverb, distortion, robotic voice. Examples: podcasts, studio voice recordings, professional voiceovers.\\
\textbf{"4" --} Studio quality with artifacts: minor noise, slight reverberation, but speech is clear and sounds very good. Example:  recordings from a microphone in a quiet room, but with background hum or light music.\\
\textbf{"3" --} Unprofessional recording: noticeable noise, distortion, poor speech clarity or the presence of any background music.  Example: recording on a cheap microphone in a room with echo, social media audio.\\
\textbf{"2" --} Very low quality: severe distortion, noise, typical telephony. Example: telephone conversation with interference, recording with loud background noise.\\
\textbf{"1" --} Low quality telephony: speech is barely understandable, intermittent sound, humming.  Example: old cell phone recording, audio from a bad VoIP call.\\
\textbf{"0" --} Non-speech: white noise, silence, unrecognizable sounds.  Example: fan noise file with no speech, broken data.

Since we posit that punctuation, stress, and phonemic annotations affect the quality of synthetic speech, we additionally collected an intonation-focused MOS (IntMOS) to capture prosodic naturalness beyond overall quality. Raters used the following rubric:\\
{\textbf{"5" --} Speech sounds like a real person in a normal conversation situation: correct stresses, logical pauses, no signs of diction or clichés.}\\
{\textbf{"4" -- } The intonation is dictation- or audiobook-like.
The speech is recognized as human speech, but the narration is similar to dictation or audiobook, there may be excessive expressiveness and some unnatural pauses, there are no errors in stresses.}\\
{\textbf{"3" -- } The intonation is indeterminate. It is difficult to tell whether it is a human or a machine: there may be individual mistakes in stresses and pauses, the structure of the intonation is broken.}\\
{\textbf{"2" -- } Mostly robotic intonation. Speech sounds synthetic, there are often errors in stresses and pauses, intonation is unfamiliar, but there are attempts to imitate a human.}\\
{\textbf{"1" -- } Clearly unnatural intonation. Intonation is clearly artificial, most of the stresses and pauses are wrong, there is no likeness to human speech.}\\
{\textbf{"0" -- } Speech is unintelligible. It is impossible to understand what is being said, intonation cannot be evaluated due to low comprehensibility or loss of meaning.}

Each audio item in our human-feedback evaluations received at least seven independent ratings from native speakers of Russian. For each clip, we aggregated ratings by taking the median. System-level scores were computed as the mean across clips, and we report 95\% confidence intervals (CI) using:

\begin{equation}
    CI = z^* \frac{S}{\sqrt{n}}
\end{equation}
where $z^*=1.96$ is the Z-score for 95\% confidence interval, $S$ is the standard deviation and $n$ is the number of clips. IntMOS is reported alongside MOS to isolate prosodic and intonational aspects that are specifically targeted by our annotation layers, providing a direct test of their contribution to perceived naturalness.
    \section{Additional Research Questions}

\subsection{RQs in Introduction}

\textbf{RQ4 (Restoration Performance):} To what extent do speech restoration models trained on data annotated with our framework outperform existing baseline models?\\
\textbf{RQ5 (Processing Throughput):} To what degree does our modular framework achieve web-scale throughput for large audio corpora?

\subsection{RQs in Experimental Setup}

\begin{table}[h!]
\footnotesize
    \caption{Real-Time Factor comparison}\label{tab:real_time_factor}
    \begin{threeparttable}
    \begin{tabular}{c c c c}
        \toprule
         Stage & Section  & RTF & hours/hour  \\ 
        \midrule
        $\texttt{Segmentation}(\cdot)$  & GPU/CPU & 0.0107 & 93.11591\\
        $ASR_{CTC}(\cdot)$  & GPU  & 0.0005 & 1741.362\\  %
        $ASR_{CTC+LM}(\cdot)$  & GPU  & 0.0007 & 1478.0087\\ %
        $ASR_{RNNT}(\cdot)$  & GPU  & 0.0062 & 160.9663\\ %
        $ASR_{VOSK}(\cdot)$  & CPU  &  0.0065 & 153.2891\\ %
        $ASR_{tone}(\cdot)$  & CPU  &  0.0128 & 77.86585\\ %
        $ROVER(\cdot)$  & CPU & 0.0002 & 5040\\
        $\texttt{Punctuator}(\cdot)$  & GPU  & 0.0003 & 3308.5851 \\ %
        $\texttt{StressPlacer}(\cdot)$  & GPU  & 0.0005 & 1879.9074\\ %
        $\texttt{G2P}(\cdot)$  & GPU  & 0.0012 & 810.7867\\
        \midrule
        PyAnnote & GPU &0.00631& 158.06722 \\%
        NISQA & GPU & 0.0003& 2857.14286 \\%
        \bottomrule
    \end{tabular}
    \begin{tablenotes}
         \item \textbf{Note:} Real-Time Factor (RTF) is defined as \(\text{processing time} / \text{audio duration}\); values $<1$ indicate faster-than-real-time processing. “hours/hour” denotes the amount of audio (in hours) processed per wall-clock hour, computed as \(1 / \text{RTF}\). The “Section” column refers to corresponding subsections in this paper where each stage is described, and “Device” indicates whether the stage runs on GPU, CPU, or both.
    \end{tablenotes}
    \end{threeparttable}
\end{table}
\begin{table*}[ht!]
\footnotesize
    \centering
    \caption{ {Speech Restoration models comparison}}\label{tab:SR}
    \begin{threeparttable}
    \begin{tabular}{ c c c c c c c c c}
         \toprule
         Model & NOI  & DIS & COL & LOU & NMOS& UTMOS & MOS $\pm$ 95\% CI & AR $\pm$ 95\% CI \\
         \midrule
         Source & 2.8884&3.4525&2.8541&3.167&2.6316&2.57&2.87$\pm$0.1&N/A\\
         SEMamba~\cite{chao2024investigation}& 3.9307&\underline{4.1226}&3.6928&3.8473&3.7824&\textbf{2.6667}&2.929 $\pm$ 0.103&0.881 $\pm$ 0.038\\
         DeepFilterNet3~\cite{schroeter2023deepfilternet3}&\underline{4.0474}&3.9886&3.5067&3.7764&3.6049&2.4665&2.137 $\pm$ 0.057&\textbf{0.964 $\pm$0.017}\\
         VoiceRestore ~\cite{kirdey2025voicerestoreflowmatchingtransformersspeech}&3.1257&3.8354&3.1551&3.4223&3.0314&2.2688&2.944 $\pm$ 0.084&0.933 $\pm$0.03\\
         MP-SENet ~\cite{lu2023mp}&3.9409&4.1193&\underline{3.7012}&\underline{3.9491}&\underline{3.8098}&\underline{2.4914}&\underline{3$\pm$0.087}&0.9234 $\pm$ 0.031\\
         MossFormer2 ~\cite{zhao2024mossformer2combiningtransformerrnnfree} &3.7777&4.0171&3.4800&3.6947&3.5138&2.3973&2.953 $\pm$0.079& 0.949$\pm$ 0.027\\
        \textbf{Balalaika (ours)}&\textbf{4.1723}&\textbf{4.1842}&\textbf{3.922}&\textbf{4.1408}&\textbf{3.8723}&2.4152&\textbf{3.2625 $\pm$ 0.116}&\underline{0.955 $\pm$ 0.039}\\
         \bottomrule
    \end{tabular}
    \begin{tablenotes}
    \item \textbf{Note:} The table shows a comparison of different speech restoration models whose weights and inference code are taken from the original implementations. The table also shows the metrics of the original degraded sample (source) and the SEMamba model trained on the first part of our dataset (ours). The following metrics were used: NISQA\cite{Mittag_Naderi_Chehadi_Möller_2021} (NOI, COL, DIS, LOU, NMOS), UTMOSv2\cite{baba2024utmosv2} (UTMOS), the manual MOS with 95\% confidence intervals (MOS $\pm$ 95\% CI), and the accent rate with 95\% confidence intervals (AR $\pm$ 95\% CI), which represents the percentage of audio that has an accent. The highest and second-highest scores for the metrics are shown in \textbf{bold} and \underline{underlined} text.
    \end{tablenotes}
    \end{threeparttable}
\end{table*}

\subsubsection{RQ4: Speech Restoration Performance}

We test whether a restoration model trained on our data outperforms strong baselines on a fixed out-of-sample benchmark, isolating dataset quality from model novelty. Baselines are evaluated with official weights and settings, unmodified, to ask if training on our data can beat off-the-shelf models on the same test audio. We train a single restoration model (SEMamba~\cite{chao2024investigation}) with Adam (lr=$5\times10^{-4}$), batch size 20, for $10^5$ steps, using MUSAN~\cite{musan2015} and  Room Impulse Responses(RIRs)~\cite{7953152} to construct degradations; we use the last checkpoint for evaluation. Training uses the entire corpus to reflect the end-to-end benefit of our approach.

The evaluation set is a 20-hour subset of SOVA RuYouTube~\cite{sova2022dataset} selected at random and held fixed for all systems; subjective metrics are computed on a uniformly sampled 200-clip subset from these 20 hours, identical across systems to enable paired comparisons. Objective quality is reported with NISQA metrics~\cite{Mittag_Naderi_Chehadi_Möller_2021} and UTMOS~\cite{baba2024utmosv2} on the full test set; subjective quality uses MOS; We also report Accent Rate (AR), defined as the proportion of clips that raters judge to carry a non-native accent, capturing artifacts that may arise when systems are trained on non-Russian or otherwise mismatched data. All baselines (SEMamba~\cite{chao2024investigation}  {(original and trained on our dataset)}, DeepFilterNet3~\cite{schroeter2023deepfilternet3}, VoiceRestore~\cite{kirdey2025voicerestoreflowmatchingtransformersspeech}, MP-SENet~\cite{lu2023mp}, MossFormer2~\cite{zhao2024mossformer2combiningtransformerrnnfree}) are run from official releases with default inference parameters on the same input audio.

\begin{table*}[ht!]
\footnotesize
    \centering
    \caption{Curated Datasets}\label{tab:datasets_comparison}
    \begin{threeparttable}
    \begin{tabular}{ c c c c c}
        \toprule
         Dataset & original dataset hours & Balalaika version hours & original link & Balalaika version link\\
        \midrule
        
        Yandex Music Podcasts & 3071 & 2165 & N/A & The link will be available at camera-ready version \\

        OpenSTT~\cite{slizhikova2019openstt} & 20108 & 431.43 & \href{https://github.com/snakers4/open_stt}{link} & The link will be available at camera-ready version\\

        ESpeech~\cite{Den4ikAI_ESpeechTechReport_2025} & 1726 & 475.892 & \href{https://huggingface.co/ESpeech}{link} & The link will be available at camera-ready version\\

        ESpeech Podcasts~\cite{Den4ikAI_ESpeechTechReport_2025} & 3200 & 900.63 & \href{https://huggingface.co/datasets/ESpeech/ESpeech-podcasts}{link} & The link will be available at camera-ready version\\

        GOLOS~\cite{karpov21_interspeech} & 1227 & 49.086 & \href{https://www.openslr.org/114/}{link} & The link will be available at camera-ready version\\

        DeepSpeech~\cite{fedoseev2017deepspeech} & 6000& 278.634 &\href{https://github.com/GeorgeFedoseev/DeepSpeech}{link} & The link will be available at camera-ready version \\






        TONE Webinars~\cite{vikhrmodels_tone_collection} & 2208& 248.91 & \href{https://huggingface.co/datasets/Vikhrmodels/ToneWebinars/}{link} & The link will be available at camera-ready version\\

        Biggest Russian Books~\cite{its5q_biggest_ru_book} & 1000& 528.247 &\href{https://huggingface.co/datasets/its5Q/biggest-ru-book}{link} & The link will be available at camera-ready version\\

        \bottomrule
    \end{tabular}
    \begin{tablenotes}
    \item \textbf{Note:} “Original dataset hours” are taken from the upstream projects or papers; “Balalaika version hours” reflect the portion processed and released within the Balalaika collection for that source, and may be updated. “Original link” points to the upstream dataset landing page; “Balalaika version link” points to the corresponding Balalaika-packaged subset with enriched annotations.  “N/A” indicates the upstream source has no single canonical landing page or redistribution is not permitted.
  \end{tablenotes}
    \end{threeparttable}
\end{table*}

\subsubsection{RQ5: Processing Throughput}

Assessing stage-wise throughput is key to understanding the scalability of our framework. Variability in source material and quality thresholds can significantly change downstream data volume, so we report per-stage metrics rather than an aggregate end-to-end figure, which would be dataset dependent and potentially misleading. Throughput is measured as Real-Time Factor defined as the total processing time divided by the total audio duration. We also report hours processed per wall-clock hour. All measurements use a single machine with 3x RTX5060Ti 16GB GPU, 2x RTX4060Ti 16GB and dual Intel Xeon E5-2699a v4 CPU, running four parallel processes and single-item batches.

\subsection{RQs from Results and Discussion}

\subsubsection{RQ4: Speech Restoration Performance}

Models trained on our data outperform baseline systems, with our SEMamba~\cite{chao2024investigation} achieving the highest human MOS and the strongest overall NISQA profile among all methods in Table \ref{tab:SR}. The objective predictors and subjective ratings are aligned for our model, and Accent Rate remains on par with leading alternatives, indicating that the gain in perceived restoration quality is consistent across metrics and does not introduce accent-related artifacts.

\subsubsection{RQ5: Processing Throughput}

Our framework processes all stages faster than real time, confirming scalability for continuous annotation. As shown in Table \ref{tab:real_time_factor}, the most demanding modules remain below the real-time threshold. This stage-wise balance prevents single-module bottlenecks and demonstrates that the modular design supports web-scale dataset creation without compromising speed or annotation depth.

    \section{Ethics}\label{sec:ethics}

\textbf{Privacy \& Consent.} This work processes publicly available audio to create annotations for generative speech research. We neither collect, infer, nor release speaker identities or demographics; single-speaker filtering excludes only overlapping speech. Released artifacts contain solely auto-generated annotations aligned to source IDs and timestamps, which risk deanonymization if combined with external data.

\textbf{Licensing.} Source material is included only when platform terms or licenses allow derivative research use, or where fair use may apply. We do not host third-party audio if prohibited, and require users to comply with local IP law.


\textbf{Misuse \& Safety.} TTS/enhancement models trained with our pipeline could be misused for impersonation, fraud, or disinformation. We recommend mitigating this risk by deploying and regularly updating audio anti-spoofing [This citations will be avalable at camera-ready version] and monitoring performance under out-of-domain conditions.

\textbf{Raters Identities \& Consent.} Manual evaluations were conducted by an anonymized pool of raters whose roles/affiliations are not disclosed; no identifiers were collected. Raters provided informed consent and participated voluntarily without coercion; no demographics were collected beyond self-attested native Russian proficiency, and no conflicts of interest were reported. The evaluated audio was not expected to contain sensitive content.

    \section{Additional Discussions}

\subsection{Annotator pool}


This study engaged 68 native Russian annotators; every audio item received at least seven independent ratings, with per-item scores aggregated by the median, and each manual metric was evaluated on a 200‑item sample; confidence intervals are reported for all metrics, only native-language status was collected about raters, and the annotation platform is [To ensure authors' anonymity, the platform name will be available at camera-ready version].

\subsection{Multilinguality}\label{appendix:discussion_multylinugality}

The pipeline is modular and can be ported beyond Russian by swapping language-dependent components, but several parts require careful adaptation to maintain accuracy and prosody fidelity across languages. Semantic VAD is the first bottleneck: SmartTurn v2~\cite{smartturnv2} currently supports a fixed set of high‑resource languages (English, Italian, French, Spanish, Dutch, Russian, German, Chinese, Korean, Portuguese, Turkish, Japanese, Polish, Hindi), so low‑resource languages would need new semantic VAD models or rule‑based fallbacks, which is outside this paper’s scope.

Quality and speaker modules are likely portable but should be validated: NISQA~\cite{Mittag_Naderi_Chehadi_Möller_2021} and PyAnnotate diarization~\cite{Bredin23} may be language-agnostic in principle, yet cross-lingual robustness must be tested before production use, especially for accents, phonotactics, and music/ad overlap patterns that differ by language and domain. 

Transcription is replaceable where strong ASRs exist~\cite{babu22_interspeech}; for low‑resource languages, choosing resilient architectures and multilingual or self‑supervised models is crucial~\cite{liang2025fieldmatters} since robust ASR remains an open challenge and directly impacts downstream punctuation and timestamp reliability.

Text enrichment layers are the most language-specific: Russian punctuation restoration models cannot be reused as-is; building or selecting per-language punctuators, or training a multilingual punctuator, remains an open research problem with variable transferability across scripts and orthographies~\cite{khare2025universal2tfrobustallneuraltext}. 

Stress placement is particularly challenging in Slavic and other movable-stress languages; beyond Russian, reliable public models are scarce, and language-specific stress predictors would need to be developed where applicable to preserve prosody cues~\cite{petrov-2025-ruaccent}.

Finally, grapheme-to-phoneme to IPA is comparatively straightforward to extend; encoder–decoder G2P models can be trained per language to supply IPA sequences for TTS and generative tasks with modest engineering effort~\cite{ylonen-2022-wiktextract}, fitting cleanly into the pipeline’s plug‑replace design.

\subsection{Future work}

First, a speech restoration module will be developed that conditions on audio, text, and auxiliary features to reconstruct high‑quality speech suitable for generative training~\cite{koizumi23_interspeech}; this will unlock currently unused recordings that were excluded by strict filtering, repurposing them as supervised or self‑supervised restoration targets to expand training diversity. 

Second, a music detector will be integrated to automatically flag and exclude segments with background music, improving data purity for both ASR alignment and downstream TTS prosody modeling in mixed‑media sources such as podcasts and YouTube excerpts. 

Third, the robustness of NISQA as a no‑reference quality estimator will be stress‑tested across domains, codecs, and accents represented, with contingency plans to calibrate or design a more resilient variant if systematic biases are observed under cross‑dataset or cross‑condition evaluation.

Fourth, an audio super‑resolution component is planned to upsample 16 kHz material to 48 kHz, targeting clearer fricatives and extended bandwidth that may improve intelligibility and timbral naturalness for modern generative vocoders and diffusion TTS models, with careful human and objective evaluation to validate perceptual benefits before large‑scale adoption.

Fifth, a speech prompt generation system will be introduced to synthesize task‑specific prompts and reference snippets, enabling controllable conditioning for style, pace, and emphasis in generative models and facilitating reproducible benchmarking of prompt‑conditioned synthesis and restoration.

    \section{Reproducibility checklist}

To support reproducibility while preserving double-blind anonymity, we provide an anonymized code repository  and release-ready dataset packaging scripts; public links to the non-anonymized repositories and hosted artifacts will be added in the camera-ready version. We release all generated annotations, but we do not attach them to the submission to avoid deanonymization via hosting metadata and repository linkage.

\begin{itemize}
    \item  Code repository: end-to-end pipeline implementation (URL omitted; public link will be provided at camera-ready).
    \item Dataset repositories: one repository per curated dataset split, containing Parquet-formatte annotation artifacts (public links will be provided at camera-ready).
    \item Non-redistributable audio sources: for Yandex Music podcasts and ESpeech podcasts, we do not redistribute audio; instead we provide a single entry-point script to (i) re-download the original audio from the upstream source and (ii) deterministically match it to our released annotations and timestamps.
    \item External model dependencies: each pipeline stage except G2P relies on external models; we provide download scripts to obtain all required weights and resources.
    \item Hosting: the packaged datasets are hosted on Hugging Face and planned to be mirrored on ModelScope; direct links will be added in the camera-ready version.
\end{itemize}

\section{Curated Data}\label{appedix:curated_data}

\subsection{Yandex Music podcasts annotated by Balalaika}

Yandex Music Podcasts serve as a large, publicly accessible catalog of Russian podcast episodes with predominantly conversational, studio-quality speech well-suited for prosody-focused TTS research; in this split, audio is not redistributed due to Russian legal restrictions, but reproducibility is ensured by providing the exact episode links, identifiers, and a \texttt{yandex\_music}-based\footnote{\url{https://pypi.org/project/yandex-music/}} Python workflow to download source audio and deterministically match it to released annotations and timestamps; no category constraints were applied during collection; for compliance, source audio may be used for research purposes only, with no redistribution and no commercial use allowed, while the released artifacts are annotations.

\subsection{OpenSTT annotated by Balalaika}

OpenSTT~\cite{slizhikova2019openstt} is a large multi-domain Russian speech corpus spanning radio, public speeches, audiobooks, YouTube, and phone calls; in the Balalaika pipeline, this source was consumed as pre-segmented clips without additional segmentation, with a simple duration filter removing all clips shorter than 3 seconds before applying Balalaika pipeline.

\subsection{ESpeech annotated by Balalaika}

ESpeech~\cite{Den4ikAI_ESpeechTechReport_2025} is a Russian speech dataset introduced in a 2025 technical report by Denis Petrov that documents the corpus composition, collection and processing pipeline, and provides an accessible PDF for reference and citation, serving as a contemporary resource accompanying related Den4ikAI speech assets and tooling.

\subsection{GOLOS annotated by Balalaika}

Golos~\cite{karpov21_interspeech} is a freely available Russian speech corpus, built from audio recorded and manually annotated via crowdsourcing, and released alongside an acoustic model and 3‑gram KenLM language models trained to facilitate benchmarking and transfer learning studies.

\subsection{DeepSpeech annotated by Balalaika}

DeepSpeech~\cite{fedoseev2017deepspeech} repository is a Russian speech fork of Mozilla’s DeepSpeech~\cite{hannun2014deepspeechscalingendtoend} that adapts the end‑to‑end architecture for Russian, provides training/config scripts and Dockerized workflows, and references data acquisition from YouTube captions alongside Russian text corpora for language modeling.

\subsection{Biggest Russian Books annotated by Balalaika}

Biggest Russian Books~\cite{its5q_biggest_ru_book} is a large multi-speaker Russian audiobook corpus on Hugging Face with almost 1000 hours of high‑quality recordings and roughly 548k utterances, packaged as WebDataset shards with per‑clip text and speaker metadata for speech research.
\else
\fi

\end{document}